\documentclass[11pt,letterpaper]{article}
\usepackage{cogsys}
\usepackage[T1]{fontenc}
\usepackage{times}
\usepackage[pdftex]{graphicx} 
\usepackage{enumitem}
\usepackage{algorithm}
\usepackage{algcompatible}

\usepackage[noend]{algpseudocode}
\usepackage{subcaption}
\usepackage{amsmath}
\usepackage{amsthm}
\usepackage{multirow}
\usepackage{hhline}

\usepackage{natbib}
\cogsysheading{X}{2021}{1-6}{9/2021}{11/2021}

\ShortHeadings{Task Modifiers for HTN Planning and Acting}
              {W.\ Yuan et al.}

\begin{document} 

\title{Task Modifiers for HTN Planning and Acting}
 
\author{Weihang Yuan\textsuperscript{1}}{wey218@lehigh.edu}
\author{Hector Munoz-Avila\textsuperscript{1}}{hem4@lehigh.edu}
\author{Venkatsampath Raja Gogineni\textsuperscript{2}}{gogineni.14@wright.edu}
\author{Sravya Kondrakunta\textsuperscript{2}}{kondrakunta.2@wright.edu} 
\author{Michael Cox\textsuperscript{2}}{michael.cox@wright.edu}
\author{Lifang He\textsuperscript{1}}{lih319@lehigh.edu}
\address{\textsuperscript{1}Department of Computer Science and Engineering, Lehigh University \\
\textsuperscript{2}Department of Computer Science and Engineering, Wright State University}
\vskip 0.2in
 
\begin{abstract}

The ability of an agent to change its objectives in response to unexpected events is desirable in dynamic environments. In order to provide this capability to hierarchical task network (HTN) planning, we propose an extension of the paradigm called task modifiers, which are functions that receive a task list and a state and produce a new task list. We focus on a particular type of problems in which planning and execution are interleaved and the ability to handle  exogenous events is crucial. To determine the efficacy of this approach, we evaluate the performance of our task modifier implementation in two environments, one of which is a simulation that differs substantially from traditional HTN domains.

\end{abstract}

\section{Introduction}
A key feature of cognitive systems is their reliance on structured representations of knowledge \citep{langley2012cognitive}. A well-known structured representation is hierarchical task networks (HTNs)  \citep{erol1994umcp}.  HTNs represent a series of tasks at different levels that explicitly encode two relations:
\begin{itemize}
\item Task-subtask: each task (unless it is at the root) has a parent task and one or more children (unless it is a leaf).
\item Order: sibling tasks that are at the root or ones that are children of the same parent have an ordered relation between them. 
\end{itemize}
Because of its expressivity, the HTN framework has been used in many applications \citep{nau2004applications} and cognitive architectures \citep{choi2018evolution,laird2019soar}.

Another recurring research topic in cognitive systems is goal reasoning, which is based on the following idea: an agent supervises its own actions in an attempt to achieve some goals in an environment; when the agent encounters discrepancies between its expectations and actual observations about the environment, new goals are generated as a result. \cite{cox2016midca} use a mechanism based on cause and effect. For example, $c \to d$ indicates a cause $c$ (e.g., there is a minefield) for a discrepancy $d$ (e.g., a mine is encountered). When the discrepancy $d$ is encountered, the goal $\neg c$ is generated (e.g., clear the minefield) to obviate the discrepancy $d$.

Goal reasoning uses the notion of goal. A goal is a set of ascertainable conditions (i.e., either true or false) in a state. For example, the condition ``the agent encounters a mine'' can be determined conclusively. In contrast, HTN planning is concerned with tasks, which may or may not be goals\footnote{Some approaches reason with goals only. For example,  hierarchical goal network (HGN) planning \citep{shivashankar2013hierarchical}, a variant of HTN, use goals in hierarchies instead of tasks. Others have combined tasks and goals \citep{nau2021gtpyhop}. }. In general, a task is symbolic: it is defined by the methods that describe how to achieve it. Hence tasks are (ambiguously) considered  ``activities'' to be performed. ``Seek nearby mines'' is an example of a task that does not represent a specific state condition and therefore it is not a goal. Despite the dichotomy between tasks and goals, there is no theoretical constraint we know of that hinders the application of goal reasoning techniques to systems that reason with tasks.

Along these lines, we introduce a modification of the HTN  planning algorithm SHOP \citep{nau1999shop}. The original algorithm and variants are widely adopted \citep{nau2004applications,cox2016midca}.  Our modification enhances HTN planning in two ways: (1)  planning and execution of actions are interleaved to handle interaction in nontraditional HTN domains; (2) the agent can change its tasks in response to discrepancies encountered during execution with the use of task modifiers, functions that receive a task list and a state and produce a new task list.

\section{Preliminaries} 

Our proposed approach is based on HTN planning, although we use more general definitions than the nomenclature used by \cite{nau1999shop} to better match our implementation.

A state variable describes an attribute of a domain world. For example, $\mathit{loc}(\mathit{agent})=(0,0)$ indicates that the agent is at location $(0,0)$. A state $s$ is a set of all state variables. The set of all states is denoted $S$.

Tasks symbolically represent activities in a domain. A task $t$ consists of a name and a list of arguments and can be either primitive or compound. The set of all tasks is denoted $T$. A task list is a sequence of tasks $\tilde{t}=(t_1,\dots,t_n)$. The set of all possible task lists is denoted $\tilde{T}$ (excluding the empty list).

A primitive task can be achieved by an action. An \textit{action} is a 2-argument function $a:S\times\{t\}\to S  \cup \{\texttt{nil}\}$,
\begin{equation}
    a(s,t) = s',
\end{equation}
where $s$ is a state and $t$ is a primitive task. If $a$ is applicable to $t$ in $s$, $s'\in S$.  Otherwise, $s' = \texttt{nil}$.

A method describes how to decompose a compound task into subtasks. A \textit{method} is a 2-argument function $m:S\times\{t\}\to \tilde{T}  \cup \{\texttt{nil}\}$,
\begin{equation*}
    m(s,t) = \tilde{t},
\end{equation*}
where $s$ is a state and $t$ is a compound task. If $m$ is applicable to $t$ in $s$, $ \tilde{t}\in \tilde{T}$.  Otherwise, $\tilde{t} = \texttt{nil}$.

An HTN planning problem is a 3-tuple $(s, \tilde{t}, D)$, where $s$ is a state, $\tilde{t}=(t_1, \dots,t_n)$ is a task list, and $D$ is the set of all actions and methods. A plan is a sequence of actions. Solutions (plans) for HTN planning problems are defined recursively.
A plan $\pi=(a_1,\dots,a_m)$ is a solution for the HTN planning problem $(s, \tilde{t}, D)$ if one of the following cases is true:
\begin{enumerate}
  \item If $\tilde{t}=\emptyset$, then $\pi=\emptyset$ (the empty plan).
  \item If $\tilde{t}\neq \emptyset$: 
    \begin{enumerate}
        \item If $t_1$ is primitive, $a_1$ is applicable (i.e., $a_1(s,t_1) \neq \texttt{nil}$), and $(a_2,\dots,a_m)$ is a solution for  $(a(s,t_1),(t_2, \dots,t_n),D)$. 
        \item If $t_1$ is compound, there is an applicable method $m(s,t_1) \neq  \texttt{nil}$, and $\pi$ is a solution for $(s, (m(s,t_1), t_2, \dots, t_n),D)$.
    \end{enumerate}
\end{enumerate}

In Case 2 (a), after applying $a_1$, the remainder plan $(a_2,\dots,a_m)$ is found to be a solution for the remaining tasks $(t_2 \dots,t_n)$. In Case 2 (b),  the compound task $t_1$ is replaced with $m(s,t_1)$, and $\pi$ is found to be a solution for the new task list $(m(s,t_1), t_2, \dots, t_n)$.

\section{Task Modifiers}

In this section, we describe an extension to HTN called task modifiers. We provide an example of a task modifier in a marinetime vehicle simulation domain. Then we describe an algorithm that integrates task modifiers and SHOP.

The motivation for task modifiers is to provide a mechanism that handles unexpected events in some domains. Notably, this type of domain has the following characteristics: 
\begin{itemize}
    \item  The agent observes an external environment and interacts with it through actions. The dynamics (state transitions) of the environment are defined by a set of Equations~(1). In contrast, traditional HTN planning domains use operators to transform states.
    \item Actions are irreversible in the sense that the environment cannot revert to a previous state by editing state variables.
    \item The agent does not have full knowledge of the environment's dynamics. After an action is applied,  the environment transitions to a new state. The agent needs to observe and acquire information about the state. This necessitates interleaved planning and execution of actions.
    \item The agent's observations are partial. The agent needs to make inferences about the values of variables not observed.
    \item The environment is episodic. A terminal signal is sent when an episode ends.
\end{itemize}

Traditional HTN planners recursively decompose high-level tasks into simpler ones. As discussed in Section~2, the  agent's task list $\tilde{t}=(t_1,\dots,t_n)$ can only be modified in one of two ways:
\begin{enumerate}
    \item If $t_1$ is primitive and an applicable action for $t_1$ exists, the new task list is $(t_2,\dots,t_n)$.
    \item If $t_1$ is compound and an applicable method $m(s,t_1)$ exists,  the new task list is \\$(m(s,t_1), t_2, \dots, t_n)$.
\end{enumerate}
Since the environment's dynamics are unknown to the agent, unexpected events may occur. For instance, the agent may encounter environmental hazards during a navigation task. This demands greater flexibility in terms of addition, deletion, and reordering of the tasks in $\tilde{t}$. For this reason, we introduce task modifiers that provide another way to modify a task list: replace $\tilde{t}$ with a new task list $\tilde{t}'$. A \textit{task modifier} is a 2-argument function $\mathit{TM}:S \times \tilde{T} \to \tilde{T}$,
\begin{equation*}
    \mathit{TM}(s,\tilde{t}) = \tilde{t}'.
\end{equation*}
The definition of task modifiers is abstract. Any procedure that receives an observation $s$ and a task list $\tilde{t}$ and outputs another task list $\tilde{t}'$ can be considered a task modifier. (When no changes are made to the task list, $\tilde{t} = \tilde{t}'$.)  An abstract task modifier is used as part of an algorithm based on SHOP in Section~3.2. The implementation of two domain-specific task modifiers are described in Section~4.2.

\subsection{Task Modifier Example}  

We show a use case of task modifiers in a marinetime vehicle simulation domain called Minefield, where the agent is tasked with maximizing the survival of some transport ships that traverse through an area. Further details of this domain are described in Section~4.1. 

\begin{figure}[h]
\centering
\includegraphics[width=.5\textwidth]{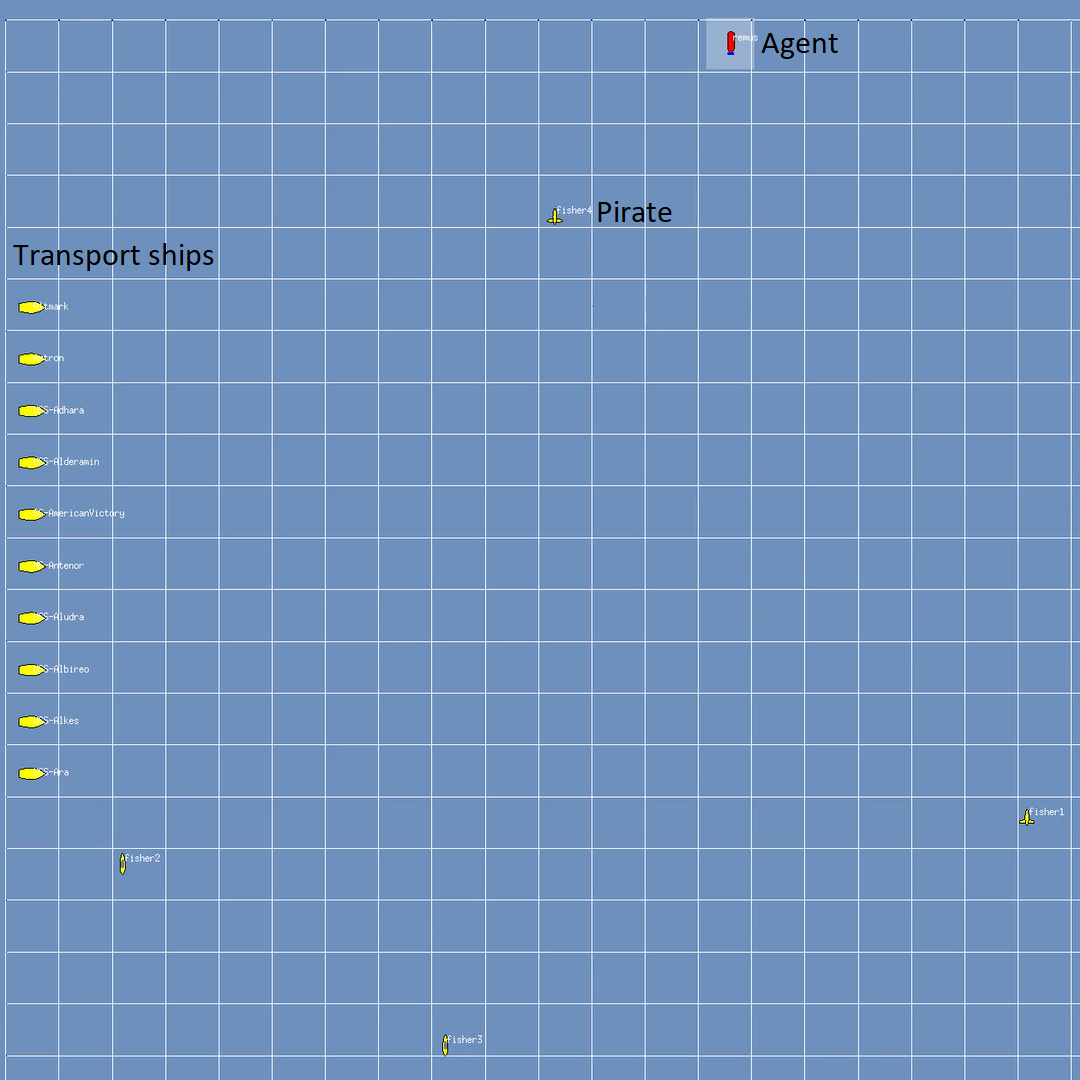}
\caption{A view of the Minefield domain at the beginning of an episode.}
\label{fig:moos}
\end{figure}

At the beginning of an episode (shown in Figure~\ref{fig:moos}), 10 transport ships are placed on the left side of the central region. The agent, the pirate boat, and three fishing boats are randomly placed in the upper and lower regions. After 20 seconds, the transport ships start to move to the right side. The pirate continuously moves to random grid cells in the central region and places mines along the way. A transport ship is destroyed when it touches a mine. The agent has no direct knowledge of which boat is the pirate. The mines in a cell are automatically cleared as the agent moves to the cell. The initial task list contains $\mathit{random\_moves}$, which keeps the agent in constant patrol of the central region.  The episode terminates when the remaining transport ships reach the right side or all the ships have been destroyed.  The objective of the agent is to maximize the number of transport ships that survive. 

The agent uses a task modifier that modifies the task list in response to unexpected events in the environment. For instance, after encountering a mine in a cell $c$, the agent decides to search nearby cells for more mines to clear:
  \begin{align*}
   (\mathit{random\_moves}) \Rightarrow  (\mathit{search\_near}(c),\mathit{random\_moves})
   \end{align*}
Alternatively, the agent may decide to pursue a suspect boat $b$ and abandon other tasks: 
  \begin{align*}
   (\mathit{search\_near}(c),\mathit{random\_moves}) \Rightarrow (\mathit{follow}(b))
   \end{align*}

To achieve a similar capability without a task modifier, each method has to be rewritten to handle dynamic events the same way a task modifier would. The code would be more complicated than just using a task modifier. Additionally, a method replaces a single task and is  agnostic of other tasks in a task list. In contrast, a task modifier is more flexible and can change any part of the task list.

\subsection{Integrating Task Modifiers and SHOP}

We now describe an algorithm that integrates SHOP with task modifiers and interleaves planning and execution. The pseudocode is shown in Algorithm~1. The type of task modifier described in the algorithm is abstract; the details of two domain-specific task modifiers are discussed in Section~4.2.

The differences between the original SHOP algorithm and our algorithm are underlined. At each time step, the agent observes the state of the environment and executes an action without full knowledge of state variables.

The algorithm starts with the procedure PLAN-ACT-TM (line 1) as an episode begins. The agent observes the current state $s$ (line 2). The procedure SEEK-PLAN-ACT-TM is called (line 3). 

SEEK-PLAN-ACT-TM (line 4) has a recursive structure similar to the HTN solution cases described in Section~2, but it does not maintains a plan because planning and execution are interleaved. If the current task list is empty or the episode terminates (line 5), the procedure returns (line 6). If the first task $t$ in the task list is primitive (line 8) and an applicable action exists (line 9), the action is executed (line 10). Then the agent observes the next state $s'$ (line 11). The task modifier receives $s'$ and the remaining tasks $R$ and updates the task list (line 12). The updated task list is passed to a recursive call of SEEK-PLAN-ACT-TM (line 13). If $t$ is compound and an applicable method $m$ is found (line 17), $t$ is replaced by its reduction (subtasks) by $m$ (line 18).  When no applicable action or method is found, the procedure returns $\texttt{nil}$ indicating a failure (lines 15 and 21).

The task modifier $\mathit{TM}$ is only called (line 12) after an action is applied (line 10) but not after a method decomposes the first task in the task list (line 18). This is a design choice based on the experimental domains, not a theoretical limitation.  After an action is executed,  unexpected changes (from the agent's perspective) may occur in the environment. In contrast, task decomposition is nearly instantaneous because the task list and the set of methods are internal to the agent.

\begin{algorithm}[!htbp]
\caption{SHOP with Task Modifier}
\label{algo_task}
\begin{algorithmic}[1]
\Procedure{plan-act-tm}{$\tilde{t},D$} 
\State \underline{observe $s$}
\State \Return \Call{seek-plan-act-tm}{$s,\tilde{t},D$}
\EndProcedure
\Statex
\Procedure{seek-plan-act-tm}{$s,\tilde{t},D$}
\If{$\tilde{t} =\emptyset$ \textbf{or} \underline{the episode terminates} } 
\State \Return $s$
\EndIf
\State $t\gets$ the first task in $\tilde{t}$; $R\gets$ the remaining tasks
\If{$t$ is primitive}
\If{there is an action  $a(s,t) \neq \texttt{nil}$ }
 \State \underline{apply $a$}
 \State \underline{observe $s'$}
 \State \underline{$R\gets \mathit{TM}(s',R)$}

 \State \Return  \Call{seek-plan-act-tm}{$s',R,D$}
 \Else \State \Return \texttt{nil}
 \EndIf
\Else
    \For{every method $m(s,t) \neq \texttt{nil}$}
        \State $s\gets$ \Call{seek-plan-act-tm}{$s,(m(s,t),R),D$}
        \If{$s \neq $ \texttt{nil}}
          \State \Return $s$
         \EndIf
    \EndFor
    \State \Return \texttt{nil}
\EndIf

\EndProcedure
\end{algorithmic}
\end{algorithm}

\section{Experiments}

To demonstrate the usage of task modifiers, we tested our implementation in two domains.\footnote{The code is available at \texttt{https://github.com/ospur/htn-tm}.} Both domains have some of the characteristics described in Section~3 that are atypical of traditional HTN domains. The intent of the experiments is to provide a qualitative comparison of our implementation and two simple baselines.

\subsection{Domains}

\paragraph{Rainy Grid.} In a $10 \times 10$ grid, the agent and a beacon randomly start at different locations and neither is at the exit, which is always in the bottom right corner. Each action produces a numerical reward. Rain occurs with a probability of $p$ and affects an action's reward. The agent knows the locations of the beacon and the exit but not $p$.  If the agent reaches the beacon, the rain stops until the end of the current episode. The episode terminates when the agent reaches the exit. The objective is to maximize the episodic cumulative reward. Rainy Grid has the following tasks:
\begin{itemize}
     \item $\mathit{move}(\mathit{dir})$. (Primitive) The agent moves one step right, up, left, or down. If it is not rainy, the action for this task has a reward of $-1$. If it is rainy, the action does nothing and has a reward of $-5$.

    \item $\mathit{go\_to}(\mathit{dest})$. (Compound) $\mathit{dest}$ is either the beacon or the exit. The method for this task recursively decomposes it into $(\mathit{move}(\mathit{dir}),\mathit{go\_to}(\mathit{dest}))$, where $\mathit{dir}$ is the direction toward $\mathit{dest}$.
\end{itemize}

\paragraph{Minefield.}\footnote{The domain was created using Mission Oriented Operating Suite Interval Programming (MOOS-IvP).} Continuing the description in Section 3.1, the entire area is a $20\times20$ grid; the central region is $20 \times 10$. The pirate continuously selects a random cell in the central region and moves toward it. At each step, with a probability of $p$, the pirate places 20 mines according to a multivariate Gaussian distribution. Minefield has the following tasks ($c$ is a grid cell and $b$ is a boat):
\begin{itemize}
 \item $\mathit{move}(c)$. (Primitive) The action for this task is applicable if the agent is adjacent to $c$. (The agent can move diagonally.) 
 \item $\mathit{arrest}(b)$. (Primitive) The action for this task is applicable if the agent is adjacent to $b$. If $b$ is the pirate, it ceases any activity for the rest of the episode. Otherwise, nothing happens.
 \item $\mathit{random\_moves}$. (Compound) This task is to randomly patrol the central region. The method for this task recursively decomposes it into $(\mathit{move\_diag}(c_1),\mathit{random\_moves})$, where $c_1$ is a random cell. 

 \item $\mathit{move\_diag}(c)$. (Compound) The task is to move along the shortest path to $c$. The method for this task recursively decomposes it into $(\mathit{move}(c_1),\mathit{move\_diag}(c))$, where $c_1$ is a cell adjacent to  the agent and in the shortest path to $c$.

 \item  $\mathit{search\_near}(c)$. (Compound) The task is to clear the mines in the adjacent cells of $c$. Note that the mines in a cell are removed once the agent reaches it.  The method for this task decomposes it into $(\mathit{move\_diag}(c_1),\dots,\mathit{move\_diag}(c_8))$, where $c_1,\dots,c_8$ are the 8 cells adjacent to $c$ in counterclockwise order.

 \item $\mathit{follow}(b)$. (Compound) The task is to follow $b$ until the agent is in the same cell as $b$.  The method for this task recursively decomposes it into $(\mathit{move}(c_1),\mathit{follow}(b))$, where $c_1$ is one step closer to the current location of $b$.
 
\end{itemize}

\subsection{Implementation of Task Modifiers}

Based on Algorithm~1, we created a modified version of Pyhop (a Python version of SHOP). Then we implement a task modifier for each domain. The following is a high-level description of the task modifiers and the intuition behind them.

\paragraph{Rainy Grid task modifier.}  The agent does not know the true probability of rain. Instead, it assumes a probability $p'$. In our experiments, we set $p'$ to 0.5. The expected cost of a single move action is computed: $E(\mathit{cost}) = \frac{1 + p'}{(1 - p')}$. Then the distances between its current location, the beacon, and the exit is computed. Multiplying $E(\mathit{cost})$ and the corresponding distance produces the expected cost of each task. The agent then decides whether to (1) directly go to the exit or (2) go to the beacon first and then the exit. The possible modifications are as follows:
\begin{enumerate}
    \item $ (\mathit{go\_to}(\mathit{exit}))  \Rightarrow (\mathit{go\_to}(\mathit{beacon}),\mathit{go\_to}(\mathit{exit}))$
    \item  $ (\mathit{go\_to}(\mathit{beacon}),\mathit{go\_to}(\mathit{exit})) \Rightarrow   (\mathit{go\_to}(\mathit{exit})) $
\end{enumerate}

\paragraph{Minefield task modifier.} Assume that the current task list is $(t_1,\dots,t_n)$. The following modifications are considered:
\begin{enumerate}
    \item When the agent encounters one or more mines in a cell $c$, the task $\mathit{search\_near}(c)$ is inserted:
  \begin{align*}
    (t_1,\dots,t_n) \Rightarrow  (\mathit{search\_near}(c),t_1,\dots,t_n)
   \end{align*}
    \item The agent estimates which boat is the pirate and decides to follow the boat $b$. This modification consists of two steps: (1) all previous $follow$ tasks are removed from the task list; (2) then a new one is added. This only triggers if the pirate has not been arrested.
     \begin{align*}
    (t_1,\dots,t_n) & \Rightarrow  (t_1',\dots,t_m') \quad \text{remove all $\mathit{follow}$ tasks} \\
      (t_1',\dots,t_m') & \Rightarrow  (\mathit{follow}(b),t_1',\dots,t_m') \quad \text{add a new $\mathit{follow}$ task} 
     \end{align*}
     \item When the agent is adjacent to a suspect boat $\mathit{b}$ after following it and the pirate has not been arrested, the task $\mathit{arrest}(b)$ is added to the task list:
       \begin{align*}
    (t_1,\dots,t_n) \Rightarrow  (\mathit{arrest}(b),t_1,\dots,t_n)
   \end{align*}
\end{enumerate}

 The identity of the pirate and the probability of it placing mines are both unknown, although the agent knows whether the pirate has been arrested. Under such conditions, the behavior of the agent is to balance between two objectives:
\begin{itemize}
\item Clear as many mines as possible: this objective impacts directly the number of transport ships that will survive. A reduced number of mines in the central region means that the transport ships are more likely to survive.
\item  Arrest the pirate as soon as possible: this objective when completed prevents the pirate from placing any more mines. However, solely focusing on this objective may cause many transport ships to be destroyed in the process.

\end{itemize}

\subsection{Baselines}

\paragraph{Rainy Grid baselines.} The purpose is to compare an agent that uses a task modifier and two baselines that have a fixed task list. The task list of Baseline~1 is $(\mathit{go\_to}(\mathit{exit}))$.  The task list of Baseline~2 is $(\mathit{go\_to}(\mathit{beacon}),\mathit{go\_to}(\mathit{exit}))$. In other words, Baseline 1 always goes to the exit directly; Baseline 2 always goes to the beacon first (turning off the rain) and then the exit.

\paragraph{Minefield baselines.} Since the domain is complex and has many variables, it is useful to establish a minimal performance baseline without any agent, the purpose of which is to show that an agent with our task modifier indeed positively impacts the survival of the transport ships. The other baseline has a random task modifier, which inserts a random task to the beginning of the task list when invoked (Algorithm~\ref{algo_task} line 12). The purpose of the random baseline is similar. It is used to show whether our task modifier is performing better than modifying the task list randomly.

\subsection{Results}

Since each domain has an objective for the agent to achieve, the performance metric is based on that objective. In the Rainy Grid domain, each move has some cost associated with it, and therefore the performance metric is the total cost. In the Minefield domain, the objective is to ensure as many transport ships survive as possible, and thus the performance metric is the number of transport ships that survive at the end of an episode.

\begin{figure}
    \centering
    \begin{subfigure}{0.45\textwidth}
        \centering
        \includegraphics[width=\textwidth]{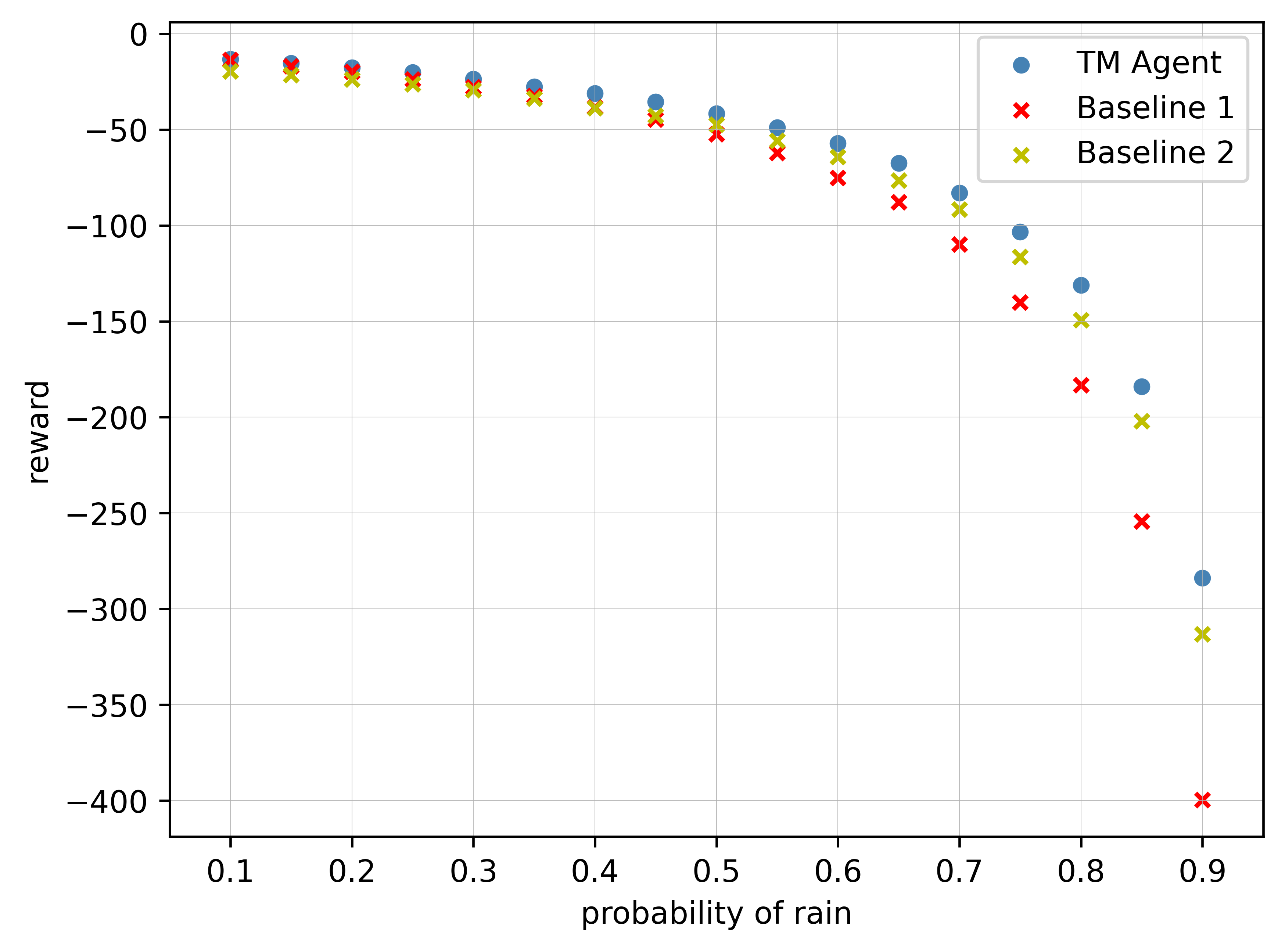}
        \caption{Rainy Grid}
    \end{subfigure}
    \begin{subfigure}{0.45\textwidth}
        \centering
        \includegraphics[width=\textwidth]{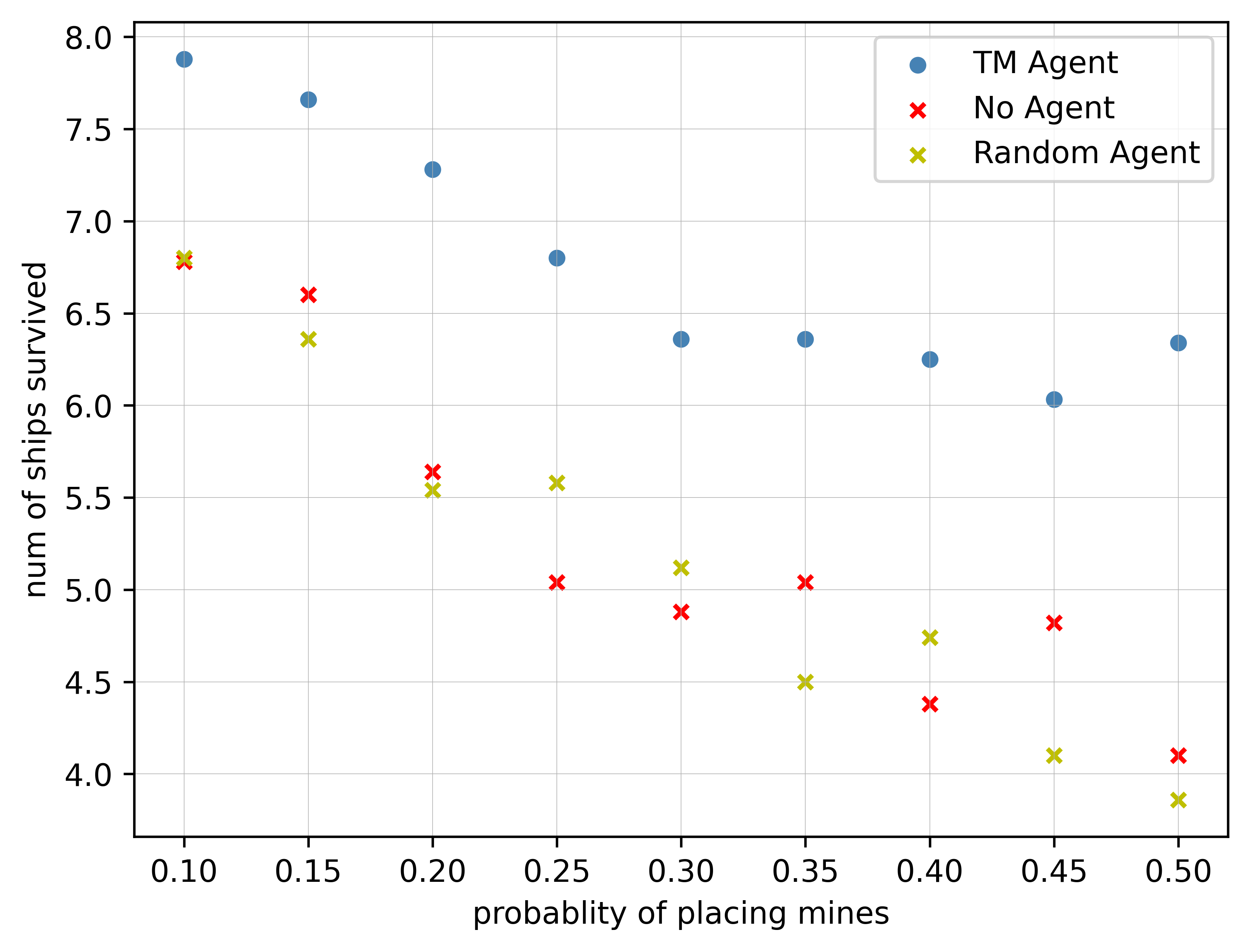}
        \caption{Minefield}
    \end{subfigure}
    \caption{Comparison of the task modifier (TM) agent and baselines.}
    \label{fig:exp}
    
\end{figure}

Figure~\ref{fig:exp} (a) shows the results of the task modifier agent and two baselines in the Rainy Grid domain. The vertical axis is the cumulative reward. The horizontal axis is the  true probability of rain. Each point is the average of 2000 runs.  When the probability of rain is low, the task modifier agent performs similarly to the baselines. As the probability increases, the task modifier agent begins to outperform the baselines.

Figure~\ref{fig:exp} (b) shows that in the Minefield domain, the task modifier agent outperforms the two baselines in all the probability configurations tested. The vertical axis is the number of transport ships that survive until the end of an episode. The horizontal axis is the probability of the pirate boat placing mines at each step. Each point is the average of 50 runs.

We perform statistical significance tests on the data from both domains. Table~\ref{tab} shows the results. The first number is the $t$-statistic and the second number is the $p$-value.  For the Rainy Grid domain,  due to the small difference in reward when the probability of rain is low, we only run $t$-tests on the data where the probability is above 0.6. We find statistically significant difference in the average reward between the task modifier agent and the baselines. For the Minefield domain, we corroborate that the task modifier agent outperforms the baselines through $t$-tests on the data where the probability of placing mines ranges from 0.2 (low) to 0.5 (high).

\begin{table}[h]
\small
\centering
\begin{tabular}{ |c|c|c|c|c| } 
\hline
\multirow{2}{*}{Rainy Grid}  &   \multicolumn{4}{|c|}{Probability of Rain}\\\cline{2-5}
  & 0.6 & 0.7 & 0.8 & 0.9 \\
\hline
TM Agent and Baseline 1 & 14.10, 4.38e-44  & 14.31, 2.46e-45 & 16.74, 7.84e-61 & 17.55, 1.83e-66 \\ 
TM Agent and Baseline 2 & 6.01, 2.01e-9 & 5.21, 2.04e-7 & 6.66, 3.05e-11 & 4.89, 1.03e-6 \\
\hline
\multirow{2}{*}{Minefield}  &   \multicolumn{4}{|c|}{Probability of Placing Mines}\\\cline{2-5}
  & 0.2 & 0.3 & 0.4 & 0.5 \\
\hline
TM Agent and No Agent  & 3.86, 0.0002 &  3.03, 0.003 & 3.97, 0.0001 & 5.10, 1.63e-6\\ 
TM Agent and Random Agent & 4.37, 3.10e-5 & 2.34, 0.021 & 3.25, 0.0016 & 5.78, 8.93e-8 \\
\hline
\end{tabular}
\caption{$t$-statistics and $p$-values of select data from the experiments.}
\label{tab}
\end{table}

\section{Related Work}

The idea of organizing tasks hierarchically was originally proposed by \cite{sacerdoti1974planning}. Works on HTN planning established the theoretical underpinnings of tasks, methods, and actions \citep{erol1994umcp,wilkins2000using}. The dominant HTN planning paradigm is ordered task decomposition as implemented in SHOP and SHOP2. It commits to a total order of the tasks unlike earlier planners \citep{wilkins2000using}, which use partial-order planning. SHOP2 allows partial order between tasks in methods, but when tasks are committed they need to be totally ordered. While losing the flexibility of partial-order plan representations, ordered task decomposition has resulted in significant running time speed gains, one of the key reasons why this paradigm has become dominant. Although our extension is based on SHOP, the same design principle works for SHOP2 as the only difference between SHOP and SHOP2 is how the first task is selected; SHOP2 maintains a partially ordered list and hence selects a task that has no predecessor \citep{nau2001total}.

HTN planning is shown to be strictly more expressive than STRIPS planning \citep{erol1994htn}. Our work is based on SHOP and has the same expressive power as an HTN planner. This type of hierarchical representation has been use by several cognitive architectures. For example, SOAR \citep{laird1990integrating} integrates hierarchical execution and learning to handle dynamic environments. It dynamically selects actions based on its production memory. When a solution is successfully found, SOAR learns the decisions that led to the solution.  ICARUS \citep{choi2018evolution} uses a teleoreactive process to learn hierarchical planning knowledge whereby gaps in planning knowledge trigger a procedure to learn targeted HTNs.

Our work is closely related to the actor view of planning \citep{ghallab2014actors},  which emphasizes the need for interleaved planning and execution. It advocates hierarchical online plan generation, as stated by \cite{ghallab2014actors}, "the actor, refine, extend, update, change, and repair its plans throughout the acting process." This builds on previous studies of interleaving planning and execution. For example, \cite{fikes1971strips} propose adding inhibitors as an execution strategy for plans to ensure that invalid actions are not executed. Cognitive architectures including SOAR, ICARUS, and MIDCA use a variety of mechanisms to identify gaps in their planning knowledge detected during execution and learn to fill those gaps. \cite{sirin2004htn} combine HTN planning and execution to generate semantic web service composition plans.  Our use of methods is identical to that of \cite{sirin2004htn}: to decompose tasks into subtasks. However, task modifiers can replace task sequences.

Task insertion \citep{xiao2020refining} is used in domains where HTN methods are incomplete. Primitive tasks are inserted to fill the gaps in HTN methods. In our work task modifiers are used in response to unexpected events in the environment, not to fulfill incomplete HTN methods.

Replanning is needed when a failure in the execution of the current plan is encountered in a state  that prevents it from continuing to execute the portion of the plan yet to be executed. For instance, \cite{fox2006plan} 
 examine  two plan completion strategies: adapting the remaining plan  or  generating a new plan from the scratch starting from the failing state. In general, adapting the remaining plan is known to be computationally harder than planning from scratch \citep{nebel1995plan}.

\cite{warfield2007adaptation} present a plan adaptation algorithm for HTN replanning. The difference between replanning and our work is that task modifiers can change tasks including the input task list whereas in replanning the input task list remains the same. 

The partially observable Markov decision process (POMDP) formulation can be used for planning when states are partially observable. The formulation enables planning in advance for any contingency the agent may encounter \citep{kaelbling1998planning}. The 3-tuple $O(s,a,o)$ indicates the probability of making an observation $o$ when executing an action $a$ in a state $s$. For instance, an agent might know the layout of a labyrinth it is navigating but not its exact location within the labyrinth. This means the agent can partially observe its whereabouts. When the agent moves forward it may encounter a red wall (i.e., an observation $o$), which indicates the agent is in any of the two possible locations that have a red wall (i.e., the probability is 0.5). Using this probabilistic information, a POMDP agent can plan for every circumstance ahead of time. However, our work and goal reasoning in general deal with potentially unforeseen and unmodeled situations (e.g., no prior knowledge of $O(s,a,o)$) and the overall goal can be changed.

\section{Conclusions}

In this paper we describe an extension of the HTN planning paradigm called task modifiers. A task modifier receives a task list and a state and produces a new task list. We describe a SHOP-based algorithm that combines task modifiers and interleaved planning and execution, which provides greater flexibility when handling unexpected events in the environment. We implemented this algorithm and two domain-specific task modifiers. Our implementation is shown to be effective in two domains: a stochastic grid environment and a marinetime vehicle simulation, where the agent is tasked with protecting transport ships. 

For future work, we want to create a task modifier procedure that is more domain-independent. In addition to being a function, this version of task modifier has access to a set of modifications. A modification consists of  a task list and a set of conditions; it is applicable to a task list given an observation if the conditions holds. Applying the modification results in a new task list.  The task modifier will select appropriate modifications based on some criteria that can be learned.

\section*{Acknowledgements} This research is supported by the Office of Naval Research grants N00014-18-1-2009 and N68335-18-C-4027 and the National Science Foundation grant 1909879.

\bibliographystyle{cogsysapa}
\bibliography{references}

\end{document}